\documentclass{article}

\usepackage{microtype}
\usepackage{graphicx}
\usepackage{subfigure}
\usepackage{booktabs} 

\usepackage{hyperref}

 \usepackage[accepted]{icml2023}

\usepackage{amsmath}
\usepackage{amssymb}
\usepackage{mathtools}
\usepackage{amsthm}

\usepackage[capitalize,noabbrev]{cleveref}

\theoremstyle{plain}

\theoremstyle{definition}

\theoremstyle{remark}

\usepackage[textsize=tiny]{todonotes}

\icmltitlerunning{Homological Neural Networks}

\begin{document}

\twocolumn[
\icmltitle{Homological Neural Networks \\
\large{A Sparse Architecture for Multivariate Complexity}}




\begin{icmlauthorlist}
\icmlauthor{Yuanrong Wang}{cs,cbt}
\icmlauthor{Antonio Briola}{cs,cbt}
\icmlauthor{Tomaso Aste}{cs,cbt,src}
\end{icmlauthorlist}

\icmlaffiliation{cs}{Department of Computer Science, University College London, London, UK}
\icmlaffiliation{cbt}{UCL Centre for Blockchain Technologies, London, UK}
\icmlaffiliation{src}{Systemic Risk Centre, London School of Economics, London, United Kingdom}

\icmlcorrespondingauthor{Tomaso Aste}{t.aste@ucl.ac.uk}

\icmlkeywords{Network Science, Complex Network, Information Filtering Network, Sparse Neural Network, Higher-order Network}

\vskip 0.3in
]



\printAffiliationsAndNotice{}  

\begin{abstract}
The rapid progress of Artificial Intelligence research came with the development of increasingly complex deep learning models, leading to growing challenges in terms of computational complexity, energy efficiency and interpretability. 
In this study, we apply advanced network-based information filtering techniques to design a novel deep neural network unit characterized by a sparse higher-order graphical architecture built over the homological structure of underlying data. 
We demonstrate its effectiveness in two application domains which are traditionally challenging for deep learning: tabular data and time series regression problems. Results demonstrate the advantages of this novel design which can tie or overcome the results of state-of-the-art machine learning and deep learning models using only a fraction of parameters. The code and the data are available at \url{https://github.com/FinancialComputingUCL/HNN}.
\end{abstract}

\section{Introduction}

Computational processes can be viewed as mapping operations from points or regions in space into points or regions in another space with different dimensionality and properties. Neural networks process information through stacked layers with different dimensions to efficiently represent the inherent structure of the underlying data. Uncovering this structure is however challenging since it is typically an unknown priori. Nevertheless, studying dependencies among variables in a dataset makes it possible to characterise the structural properties of the data and shape ad-hoc deep learning architectures on it.
Specifically, the basic operation in deep neural networks consists of aggregating input signals into one output. This operation is most effective in scenarios where the spatial organization of the variables is a good proxy for dependency. However, in several real-world complex systems, modelling dependency structures requires the usage of a complex network representation. Graph Neural Networks have been introduced as one possible way to address this issue \citep{Samek2021ExplainingDN}. However, they present two main limits: (i) they are designed for data defined on nodes of a graph \cite{yang2022simplicial}, and (ii) they usually only explicitly consider low-order interactions as geometric priors (edges connecting two nodes), ignoring higher-order relations (triangles, tetrahedra, $\dots$). Instead, dependency is not simply a bi-variate relation between couples of variables and involves groups of variables with complex aggregation laws. 

In this work, we propose a novel deep learning  architecture that keeps into account higher-order interactions in the dependency structure as topological priors. Higher-order graphs are networks that connect not only vertices with edges (i.e. low-order 1-dimensional simplexes) but also higher-order simplexes \cite{Torres2020SimplicialCH}. Indeed, any higher-order component can be described as a combination of lower-order components (i.e. edges connecting two vertices, triangles connecting three edges, $\dots$). The study of networks in terms of the relationship between structures at different dimensionality is a form of homology.  
In this work, we propose a novel multi-layer deep learning unit capable of fully representing the homological structure of data and we name it Homological Neural Network (HNN). This is a feed-forward unit where the first layer represents the vertices, the second the edges, the third the triangles, and so on. Each layer connects with the next homological level accordingly to the network's topology representing dependency structures of the underlying input dataset. Information only flows between connected structures at different order levels, and homological computations are thus obtained. Neurons in each layer have a residual connection to a post-processing readout unit. HNN's weights are updated through backward propagation using a standard gradient descent approach. Given the higher-order representation of the dependency structure in the data, this unit should provide better computational performances than those of fully connected multi-layer architectures. Furthermore, given the network representation's intrinsic sparsity, this unit should be computationally more efficient, and results should be more intuitive to interpret. We test these hypotheses by evaluating the HNN unit on two application domains traditionally challenging for deep learning models: tabular data and time series regression problems.

This work builds upon a vast literature concerning complex network representation of data dependency structures \cite{CostaSantos2011ComplexityAC, Moyano2017LearningNR}. 
Networks are excellent tools for representing complex systems both mathematically and visually, they can be used for both qualitatively describing the system and quantitatively modeling the system properties. 
A dense graph with everything connected with everything else (complete graph) does not carry any information, conversely, too sparse representations are oversimplifications of the important relations. There is a growing recognition that, in most practical cases, a good representation is provided by structures that are locally dense and globally sparse.
In this paper we use a family of network representations, named Information Filtering Networks (IFNs), that have been proven to be particularly useful in data-driven modeling \cite{tumminello2005tool, barfuss2016parsimonious, Briola2023TopologicalFS}. 
The proposed methodology exploits the power of a specific class of IFNs, namely the Triangulated Maximally Filtered Graph (TMFG), which is a maximally planar chordal graph with a clique-three structure made of tetrahedra \cite{Massara2017NetworkFF}. 
The TMFG is a good compromise between sparsity and density and it is computationally efficient to construct. 
It has the further advantage of being chordal (every cycle of four or more vertices has a chord) which makes it possible to directly implement probabilistic graphical modeling on its structure \cite{barfuss2016parsimonious}. 

The rest of the paper is organised as follows. We first review, in Section~\ref{sec:litrev}, the relevant literature. Then, in Section~\ref{sec:hnn}, we introduce a novel representation for higher-order networks, the founding stone of HNNs. The design of  HNN as a modular unit of a deep learning architecture is discussed in Section~\ref{sec:modular}. Application of HNN-based architectures to tabular data experiment on Penn Machine Learning Benchmark and to multivariate time-series on solar-energy power and exchange-rates datasets are discussed in Section~\ref{sec:expTab} and  Section~\ref{sec:expTS}. Conclusions are provided in Section~\ref{sec:conclusion}.

\section{Background literature}\label{sec:litrev}

\subsection{Information Filtering Networks}

The construction of sparse network representations of complex datasets has been a very active research domain during the last two decades. There are various methodologies and possibilities to associate data with network representations. The overall idea is that in a complex dataset, each variable is  represented by a vertex in the network, and the interaction between variables is associated with the network structure.  Normally  such a network representation is constructed from correlations or (non-linear) dependency measures (i.e. mutual information) and the network is constructed in such a way as to retain the largest significant dependency  in its interconnection structure.  These networks are known as Information Filtering Networks (IFN) with one of the  best-known examples being the Minimum Spanning Tree (MST)  \cite{nevsetvril2001otakar} built from pure correlations \cite{Mantegna1998HierarchicalSI}.  The MST has the advantage of being the sparsest connected network and of being the exact solution for some optimization problems \cite{CompNet10,CompNet11}.
However, other IFNs based on richer topological embeddings, such as planar graphs \cite{tumminello2005tool,CompNet4} or clique trees and forests \cite{Massara2017NetworkFF, Massara2019LearningCF}, can extract more valuable information and better represent the complexity of the data. These network constructions have been employed across diverse research domains from finance \cite{barfuss2016parsimonious} to brain studies \cite{CompNet8}, and psychology \cite{Christensen2016PrevalenceAC}. In this paper, we use the Triangulated Maximally Filtered Graph (TMFG) \cite{Massara2017NetworkFF}, which is a planar and chordal IFN. It has the property of being computationally efficient and it can yield a sparse precision matrix with the structure of the network, thereby being a tool for $L_0$-norm topological regularization in multivariate probabilistic models \cite{aste2020topological}.

\subsection{Sparse neural networks}
Recent advances in artificial intelligence have exacerbated the challenges related to models' computational and energy efficiency. To mitigate these issues, researchers have proposed new architectures characterized by fewer parameters and sparse structures. Some of them have successfully reduced the complexity of very large models to drastically improve efficiency with negligible performance degradation \cite{Ye2018RethinkingTS, Molchanov2016PruningCN, Lee2018SNIPSN, Yu2017NISPPN, Anwar2015StructuredPO, Molchanov2017VariationalDS, Zhuo2018SCSPSC, Wang2018ExploringLR}. Others have not only simplified the architectures but also enhanced models' efficacy, further demonstrating that fewer parameters yield better model generalization  \cite{Wu2020ConnectingTD, Wen2016LearningSS, Liu2015SparseCN, Liu2017LearningEC, Hu2016NetworkTA, Zhuang2018DiscriminationawareCP, Peng2019CollaborativeCP, Louizos2017LearningSN}.

Nonetheless, in the majority of literature, sparse topological connectivity is pursued either after the training phase, which bears benefits only during the inference phase, or during the back-propagation phase which usually adds complexity and run-time to the training. A very first attempt to solve these issues is represented by network-inspired pruning methods incorporated pruning at the earliest stage of the building process, allowing for the establishment of a foundational topological architecture that can then be elaborated upon \cite{Stanley2002EvolvingNN, Hausknecht2014ANA, Mocanu2017ScalableTO}. However, the most interesting solution is represented by Simplicial NNs \cite{ebli2020simplicial} and Simplicial CNNs \cite{yang2022simplicial}. Indeed, these architectures constitute the very first attempt to exploit the topological properties of sparse graph representations to capture higher-order data relationships. Despite their novelty, the design of these neural network architectures limits them to pre-designed network data, without the possibility to easily scale to more general data types (e.g., tabular data and time series).

In this paper, we incorporate topological constraints within the design phase of the network architecture, generating a more intricate sparse topology derived from IFNs \cite{Briola2022AnatomyOA, Briola2022DependencySI, Briola2023TopologicalFS, VidalTomas2023FtxsDA}.

\subsection{Deep Learning models for tabular data}
Throughout the previous ten years, conventional machine learning algorithms, exemplified by gradient-boosted decision trees (GBDT) \cite{chen2016xgboost}, have predominantly governed the landscape of tabular data modelling, exhibiting superior efficacy compared to deep learning methodologies. Although the encouraging results presented in the literature \cite{shwartz2018representation, poggio2020theoretical, piran2020dual}, deep learning tends to encounter significant hurdles when implemented on tabular data. The works of \cite{Arik2019TabNetAI} and \cite{Hollmann2022TabPFNAT} claim to achieve comparable results to tree models, but they are all very large attention/transformer-based models. Indeed, tabular data manifest a range of peculiar issues such as non-locality, data sparsity, heterogeneity in feature types, and an absence of a priori knowledge about underlying dependency structures. Therefore, tree ensemble methodologies, such as XGBoost, are still deemed as the optimal choice for tackling real-world tabular data related tasks \cite{friedman2001greedy, prokhorenkova2018catboost, Grinsztajn2022WhyDT}. In this work, we propose a much more efficient sparse deep-learning model with similar results.

\subsection{Deep Learning models for multivariate time-series}
Existing research in multivariate time series forecasting can be broadly divided into two primary categories: statistical methods and deep learning-based methods. Statistical approaches usually assume linear correlations among variables (i.e., time series) and use their lagged dependency to forecast through a regression, as exemplified by the vector auto-regressive model (VAR) \cite{Zivot2003VectorAM} and Gaussian process model (GP) \cite{Roberts2012GaussianPF}. In contrast,  deep learning-based methods, such as LSTNet \cite{Lai2017ModelingLA} and TPA-LSTM \cite{Shih2018TemporalPA}, utilize Convolutional Neural Networks (CNN) to identify spatial dependencies among variables and combine them with Long Short-Term Memory (LSTM) networks to process the temporal information. Despite they have been widely applied across various application domains, including finance \cite{Lu2020ACM} and weather data \cite{Wan2019MultivariateTC}, these architectures do not explicitly model dependency structures among variables, being strongly limited on the interpretability side.

Recently, spatio-temporal graph neural networks (STGNNs) \cite{Shao2022PretrainingES, Shao2022DecoupledDS} have attracted interest reaching state-of-the-art performances, as exemplified by MTGNN \cite{Wu2020ConnectingTD}. STGNNs integrate graph convolutional networks and sequential recurrent models, with the former addressing non-Euclidean dependencies among variables and the latter capturing temporal patterns. The inclusion of advanced convolutional or aggregational layers accounting for sparsity and higher-order interactions has resulted in further improvements of STGNNs \cite{Wang2022NetworkFO, Calandriello2018ImprovedLG, Chakeri2016SpectralSI, Rong2020DropEdgeTD, Hasanzadeh2020BayesianGN, Zheng2020GMANAG, Luo2021LearningTD, Kim2021HowTF}. In this paper, we use the HNN unit as an advanced aggregational module to extract the dependency structure of variables from the temporal signals generated from LSTMs.

\section{A novel representation for higher order networks and its use for HNN construction} \label{sec:hnn}
The representation of undirected graphs explicitly accounts for the vertices and their connections through edges and, instead, does not explicitly account for other, higher-order, structures such as triangles, tetrahedra, and, in general, $d$-dimensional simplexes. Indeed, usually, an {undirected graph} is represented as a pair of sets, $\mathcal G = (V,E)$: the  { vertex set} $V=(v_1,...,v_p)$  and the { edge set} $E$ which is made of pairs of edges $(v_i,v_j)$. 
The associated graphical representation is a network where vertices,  represented as points, are connected through edges, represented as segments. 
This encoding of the structure accounts only for the edges skeleton of the network. However, in many real-world scenarios, higher-order sub-structures are crucial for the functional properties of the network and it is therefore convenient -- and sometimes essential -- to use a representation that accounts for them explicitly. 

\begin{figure}[h!]
    \center
    \includegraphics[width=0.45\textwidth]{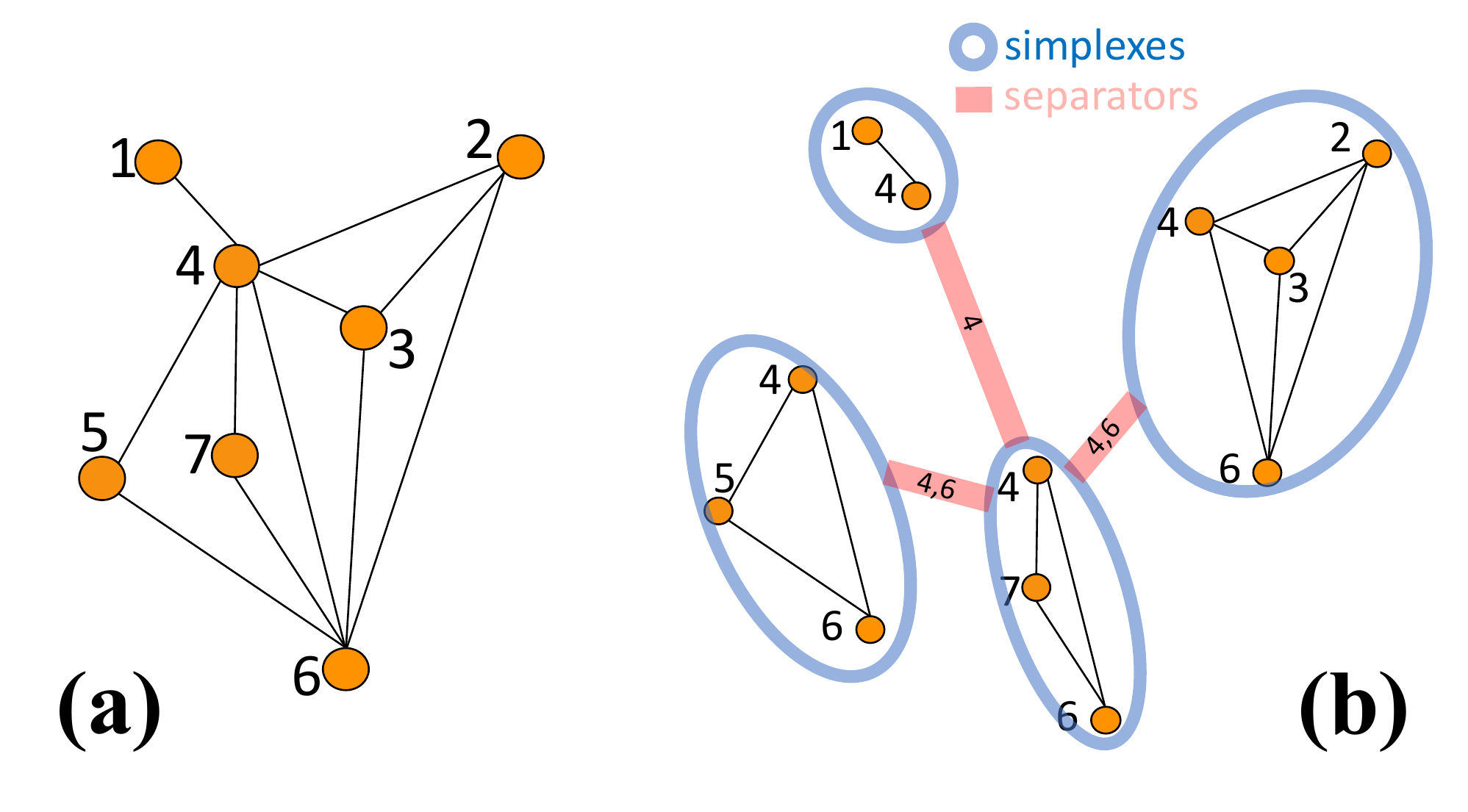}
    \caption{\footnotesize
    \label{f.CliqueTree}
    {\bf (a)} Visual example of a higher order network made of 7 vertices, 11 edges, 6 triangles, and 1 tetrahedron. 
    {\bf (b)} This higher-order network is a clique tree made of four cliques (maximal cliques highlighted in the circles) connected through three separators (the tick red edges). 
    One can observe that the separator constituted by the vertex `4' has multiplicity 1,  while the separator constituted of the edge '4-6' has multiplicity 2 and indeed it appears twice. 
     }
\end{figure}

A simple higher-order representation can be obtained by adding triplets (triangles), quadruplets (tetrahedra), etc. to the sets in $\mathcal G$. However, the associated higher-order network is hard to handle both visually and computationally. 
In this paper, we propose  an alternative approach, which consists of a {layered representation} that explicitly takes into account the higher order sub-structures and their interconnections. Such a representation is very simple, highly intuitive, of practical applicability as computational architecture, and, to the best of our knowledge, it has never been proposed before.

The proposed methodology is entirely based on a special class of networks: {chordal graphs}. These networks are constituted only of cliques organized in a higher order tree-like structure (also referred to as { `clique tree'}). This class of networks is very broad and it has many useful applications, in particular for probabilistic modeling \cite{aste2020topological}. 
A visual example of a higher-order chordal network (a clique-tree), with 7 vertices, 11 edges, 6 triangles, and 1 tetrahedron, is provided in Figure \ref{f.CliqueTree}.
In the figure, the { maximal cliques} (largest fully-connected subgraphs) are highlighted and reported, in the right panel, as clique-tree nodes. 
Such nodes are connected to each other with links that are sub-cliques called { separators}. Separators have the property that, if removed from the network, they disconnect it into a number of components equal to the {multiplicity of the separator} minus one. 
In higher-order networks, cliques are the edge skeletons of { simplexes}. A 2-clique is a 1-dimensional simplex (an edge);  3-clique is a 2-dimensional simplex (a triangle); and so on with $(d+1)$-cliques being the skeleton of $d$-dimensional simplexes. 

\begin{figure}[h!]
\center
\includegraphics[width=0.45\textwidth]{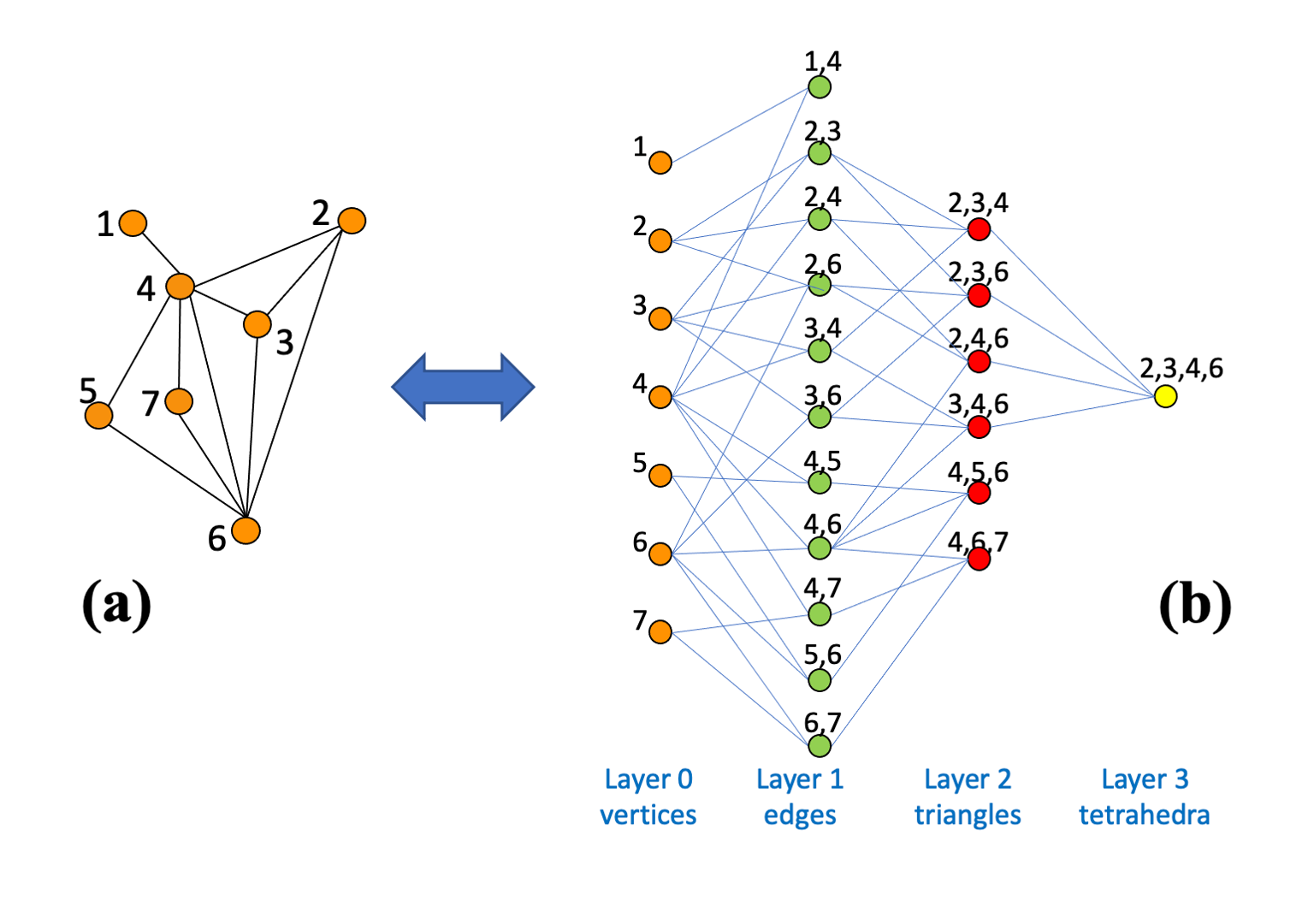}
\caption{
\label{f.Representation}
\footnotesize
Higher order homological representation of the chordal graph in Figure \ref{f.CliqueTree} (reproduced in {\bf (a)}). 
{\bf (b)} Nodes in each layer, $L_d$, represent the $d$-dimensional simplexes in the structure.
The links between nodes in layers $d$ and $d+1$ are the connections between $d$ to $d+1$ simplexes in the network. 
The degree on the left of nodes in $L_d$ is always equal to $d$.
The degree on the right of nodes in $L_d$ can instead vary. 
The $d$-dimensional simplexes with no connections towards $d+1$ are the maximal cliques in the network (i.e. the nodes in the clique tree in Figure \ref{f.CliqueTree}(b)).
}
\end{figure}

To represent the complexity of a higher-order network, we propose to adopt a layered structure (i.e. the Hasse diagram) where nodes in layer $d$ represent $d$-dimensional simplexes. The structures start with the vertices in layer 0; then a couple of vertices connect to edges represented in layer 1; edges connect to triangles in layer 2; triangles connect into tetrahedra in layer 3, and so on. This is illustrated in Figure \ref{f.Representation}. Such representation has a one-to-one correspondence with the original network but shows explicitly the simplexes and sub-simplexes and their interconnection in the structure. 
All information about the network at all dimensions is explicitly encoded in this representation including elements such as maximal cliques, separators, and their multiplicity (see caption of Figure \ref{f.Representation}). 

\begin{figure}[h!]
\center
\includegraphics[width=0.45\textwidth]{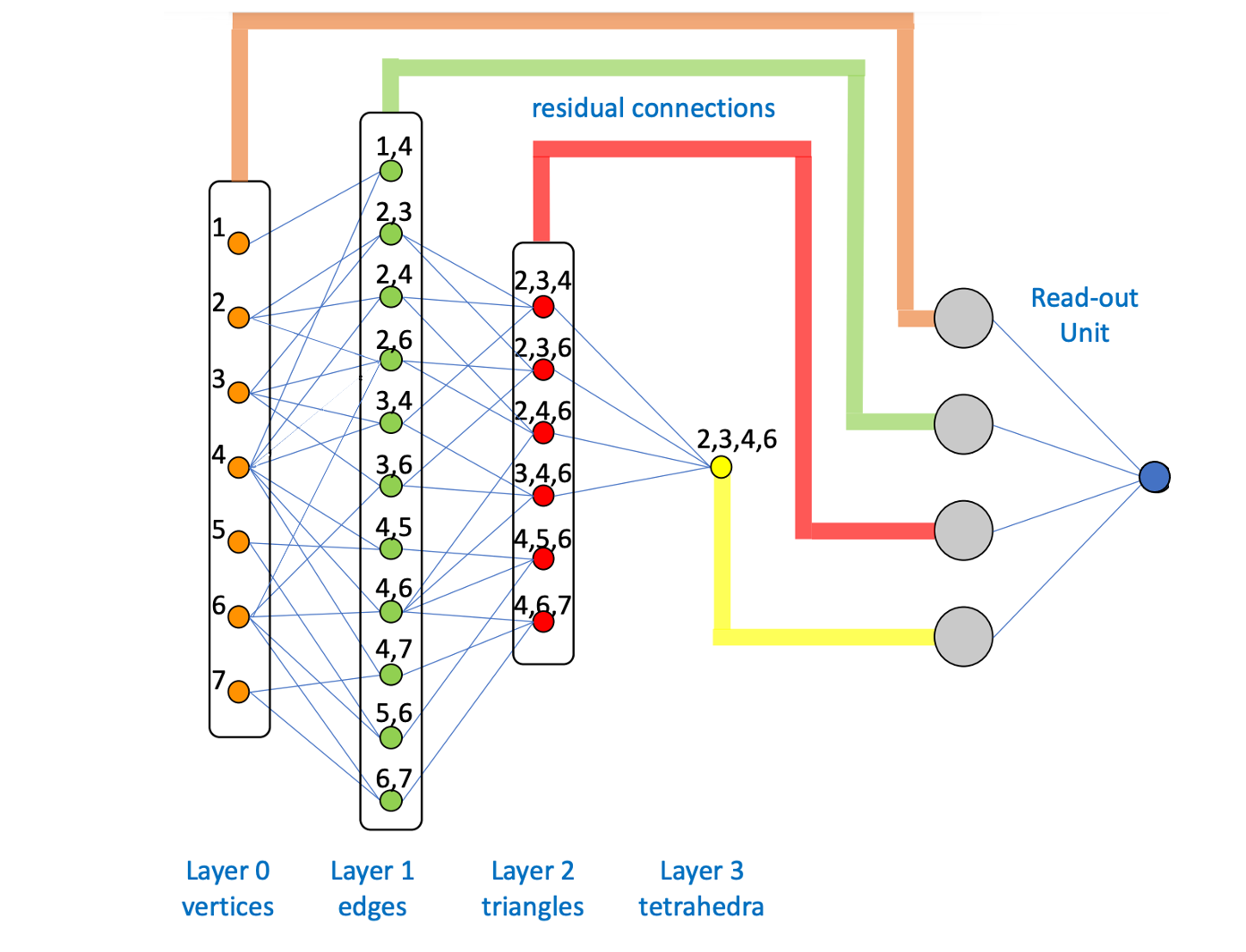}
\caption{
\label{f.HNN}
\footnotesize
The Homological Neural Network (HNN) unit is constructed by using as input layer 0 of the homological representation of the dependency structure (see Figure \ref{f.Representation}(b)) and then feeding forward through the homological layers. The output is produced by a readout unit that connects all neurons in the layers. The HNN is essentially a sparse MLP unit with residual connections. 
}
\end{figure}

It is worth noting the resemblance of this layered structure with the layered architecture of deep neural networks. Indeed, we leverage this novel higher-order network representation as the neural network architecture of the HNN unit. In our experiments, the HNN is implemented from the TMFG generated from correlations. TMFG is computationally efficient, and can thus be used to dynamically re-configure the HNN according to changeable system conditions \citep{Wang2022NetworkFO}. The HNN architecture is illustrated in Figure \ref{f.HNN}. Essentially it is made by the layered representation of Figure \ref{f.Representation} with the addition of the residual connections linking each neuron in each simplex layer to a final read-out layer. Such HNN is a sparse MLP-like neural network with extra residual connections and it can be employed as a modular unit. 
It can directly replace fully connected MLP layers in several neural network architectures. In this paper, the HNN unit is implemented using the standard PyTorch deep learning framework, while the sparse connection between layers is obtained thorugh the “sparselinear”\footnote{\url{https://github.com/hyeon95y/SparseLinear}} PyTorch library. 

\section{Design of neural network architectures with HNN units for tabular data and time series studies} \label{sec:modular}

\begin{figure}[h!]
\center
\includegraphics[width=0.45\textwidth]{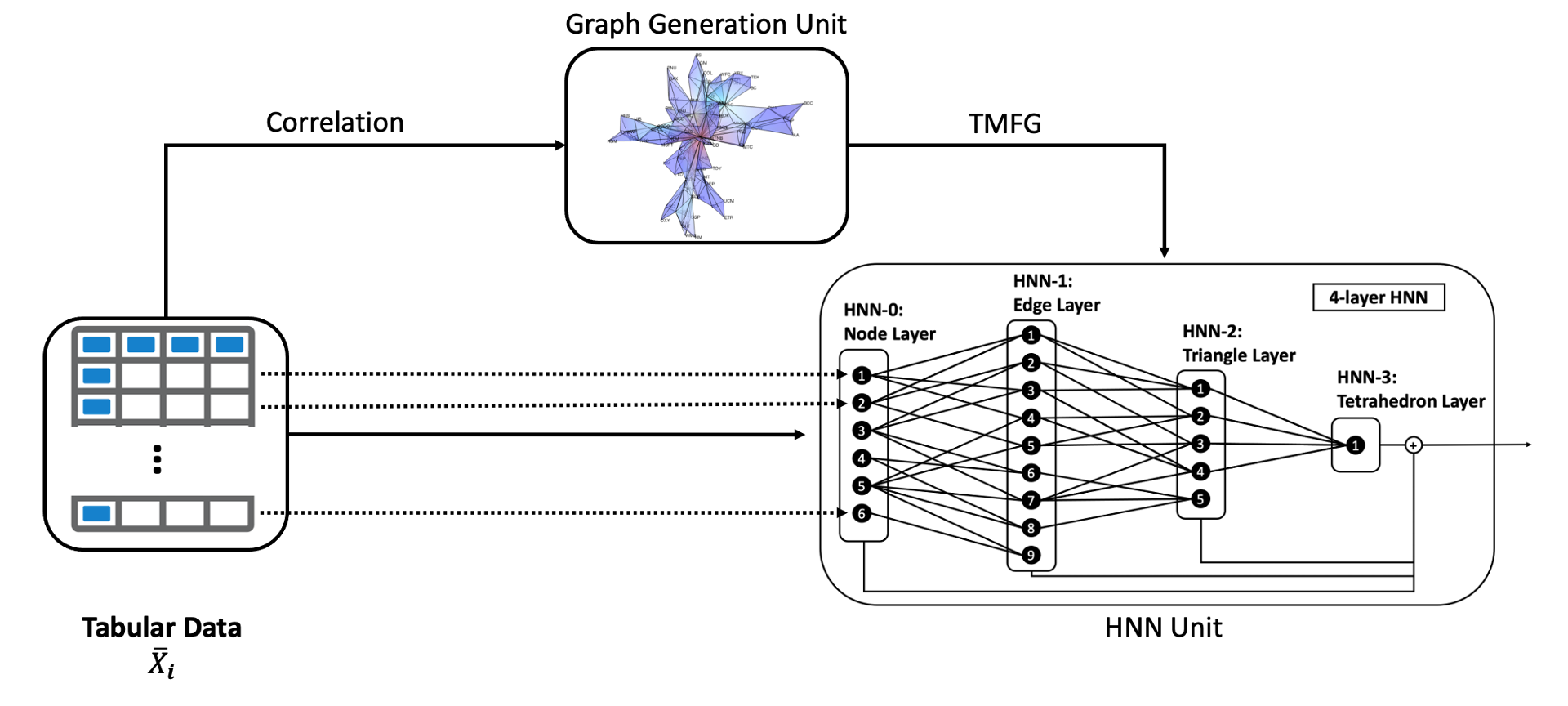}
\caption{HNN architecture for tabular data. The tabular data is processed by a Graph Generation Unit to construct a prior sparse graph to represent spatial interdependencies between the feature columns. The prior graph guides the design of the HNN unit which then processes and transforms the feature columns into the final output.
}\label{fig:tab-HNN}
\end{figure}

We investigate the performances of HNN units in two traditionally challenging application domains for deep learning: tabular data and time series regression problems. To process tabular data, the HNN unit can be directly fed with the data and it can be constructed from correlations by using the TMFG. In this case, the  HNN unit acts as a sparsified MLP. This architecture is schematically shown in Figure \ref{fig:tab-HNN}. Instead, in spatio-temporal neural networks, the temporal layers are responsible for handling temporal patterns of individual series, whereas the spatial component learns their dependency structures. Consequently, the temporal part is usually modeled through the usage of recurrent neural networks (e.g. RNNs, GRUs, LSTMs), while the spatial component employs convolutional layers (e.g. CNNs) or aggregation functions (e.g. MLPs, GNNs).

\begin{figure}[h!]
\center
\includegraphics[width=0.45\textwidth]{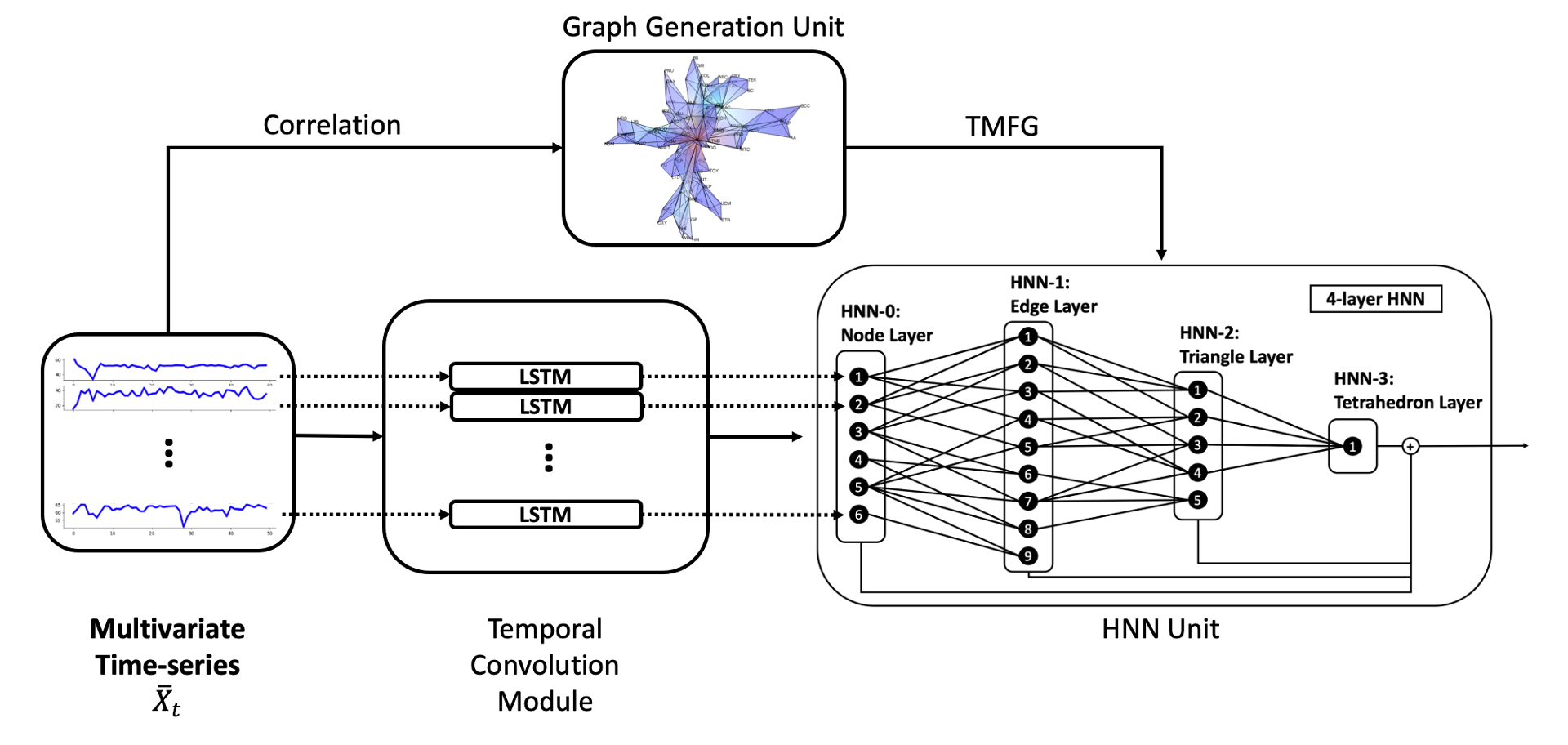}
\caption{LSTM-HNN architecture for time-series data. The multivariate time-series is processed by a Graph Generation Unit to construct a prior sparse graph to represent spatial interdependencies, and each of the multivariate time series is processed separately by LSTM in the Temporal Convolution Module to harness the temporal information. The prior graph guides the design of the HNN unit which then aggregates the single temporal representations from LSTMs into the final output.
}\label{fig:LSTM-HNN}
\end{figure}

Figure \ref{fig:LSTM-HNN} presents the spatio-temporal neural network architecture employed in our multivariate time series experiments. The architecture consists of an LSTM for the temporal encoding of each time series and a graph generation unit that takes into account the correlation between different time series. This unit models time series as nodes and pairwise correlations as edges by imposing the topological constraints typical of the TMFG: planarity and chordality. The HNN is built based on the resulting sparse TMFG and aggregates each of the encoded time series from the LSTM, generating the final output.


\section{Results}\label{sec:exp}

\subsection{Tabular Data}\label{sec:expTab}


We test performances of the HNN architecture on the Penn Machine Learning Benchmark (PMLB) \citep{romano2021pmlb} regression dataset. We select datasets with more than 10,000 samples, and we split each of them into a 70\% training and 30\% testing set.

The R2 scores for the HNN architecture are reported in Table \ref{Tab:pmlb_reg_table1} for groups of PMLB regression datasets with different number of variables. We first compare HNN performances with the ones achieved by a Multi-Layer Perceptron (MLP) with same depth (i.e. 4 as imposed by the TMFG in the HNN's building process). We also test a sparse MLP (MLP-HNN) with the same sparse structure of the HNN but without the residual connections from each layer to the final read-out layer, and a standard MLP with residual connections to the final read-out layer, MLP-res. All these architectures are optimized using gradient descent for the parameters and grid search for the hyper-parameters.
It is evident from Table \ref{Tab:pmlb_reg_table1} that the HNN architecture largely outperforms the other neural network models. It is also evident that the homological structure of HNN is the main factor associated with improved performances (indeed MLP-HNN outperforms MLP) while the residual connections between layers are not the factor that makes HNN best performing (indeed MLP-res does not outperforms HNN).

\begin{table*}[h!]
\center

\begin{tabular}{l|llll|llll|llll}
\toprule
{} &     
\multicolumn{4}{|c}{\# variable $\in [0, 20)$} &
\multicolumn{4}{|c|}{\# variable $\in [20, 40)$} & 
\multicolumn{4}{|c}{\# variable $ > 40$} 
\\
\midrule
{} &    mean &   10th &  50th  &  90th  &     mean &   10th  &  50th  &  90th  &     mean &   10th  &  50th  &  90th  \\
\midrule
\midrule
HNN    &  ${\bf0.70}^{***}$ &   0.45 &  0.78 &  0.93 &  ${\bf0.75}^{**}$ &   0.55 &  0.78 &  0.91 &   ${\bf 0.89}^{**}$ &  0.78 &  0.92 &  0.96 \\
\midrule
\midrule
MLP-HNN  & -9.64 &  -8.97 &  0.01 &  0.82 &  0.21 &  -0.01 &  0.03 &  0.56 &  0.55 &  0.14 &  0.54 &  0.96 \\
\midrule
MLP-res & -5.14 &   0.02 &  0.79 &  0.94 &  0.40 &   0.01 &  0.27 &  0.84 &  0.32 &  0.01 &  0.19 &  0.75 \\
\midrule
MLP    & -7.18 &  -0.63 &  0.19 &  0.87 &  0.09 &  -0.01 & -0.00 &  0.22 &  0.12 & -0.14 & -0.00 &  0.50 \\

\bottomrule
\end{tabular}

\vspace{0.5pt}
    \caption{R2 score from different models on PMLB regression dataset, with different number of variables. The best-performing average result is highlighted in bold, and ${}^{*}$ denotes $1\%$ significance, ${}^{**}$ for $0.1\%$ and ${}^{***}$ for $0.001\%$ respectively from paired T-test of the second best performing model result against HNN result. 
    We also report the 10\% 50\% and 90\% quantiles.
    }
    \label{Tab:pmlb_reg_table1}
\end{table*}

It is commonly acknowledged that neural network models do not perform well on tabular data \cite{Borisov2021DeepNN}; tree-based models and the gradient boosting framework represent instead the state-of-the-art \cite{ShwartzZiv2022TabularDD, Grinsztajn2022WhyDT}. We therefore compare the HNN results with baseline models including Linear Regression (LM), Random Forest (RF), Light Gradient Boosting Machine (LGBM), Extreme Gradient Boosting Machine (XGB) \cite{Ke2017LightGBMAH, chen2016xgboost}. The experiments are performed on the same datasets using the optimization and tuning pipeline descirbed above.

\begin{table*}[h!]
\center
\begin{tabular}{l|llll|llll|llll}
\toprule
{} &     
\multicolumn{4}{|c|}{\# variable $\in [0, 20)$} &
\multicolumn{4}{|c|}{\# variable $\in [20, 40)$} & 
\multicolumn{4}{|c}{\# variable $> 40$} 
\\
\midrule
{} &    mean &   10th &  50th  &  90th  &     mean &   10th  &  50th  &  90th  &     mean &   10th  &  50th  &  90th  \\
\midrule
\midrule
HNN    &  0.70 &   0.45 &  0.78 &  0.93 &  0.75 &   0.55 &  0.78 &  0.91 &  0.89 &  0.78 &  0.92 &  0.96 \\
\midrule
\midrule
XGB    &  ${\bf0.80}^{*}$ &   0.52 &  0.85 &  0.95 &  {\bf0.91} &   0.83 &  0.92 &  0.98 &  {\bf0.92} &  0.88 &  0.92 &  0.96 \\
\midrule
LGBM   &  0.65 &   0.00 &  0.81 &  0.95 &  0.89 &   0.81 &  0.91 &  0.97 &  0.91 &  0.87 &  0.91 &  0.95 \\
\midrule
RF     &  0.78 &   0.47 &  0.85 &  0.98 &  0.87 &   0.76 &  0.89 &  0.98 &  0.89 &  0.85 &  0.87 &  0.95 \\
\midrule
LM     &  0.53 &   0.12 &  0.64 &  0.90 &  0.36 &   0.01 &  0.28 &  0.74 &  0.34 &  0.11 &  0.25 &  0.65 \\

\bottomrule
\end{tabular}

\vspace{0.5pt}
    \caption{R2 score from different models on PMLB regression dataset, with a number of variables between 0 and 20, between 20 and 40, and larger than 40. The best-performing average result is highlighted in bold, and ${}^{*}$ denotes $1\%$ significance from paired T-test of the best-performing model result against HNN result. We also report the 10\% 50\% and 90\% quantiles. 
    The absence of ${}^{*}$ indicates statistical equivalence between the best-performing and HNN.
    }
    \label{Tab:pmlb_reg_table2}
\end{table*}

\begin{figure}[h!]
\center
\includegraphics[width=0.45\textwidth]{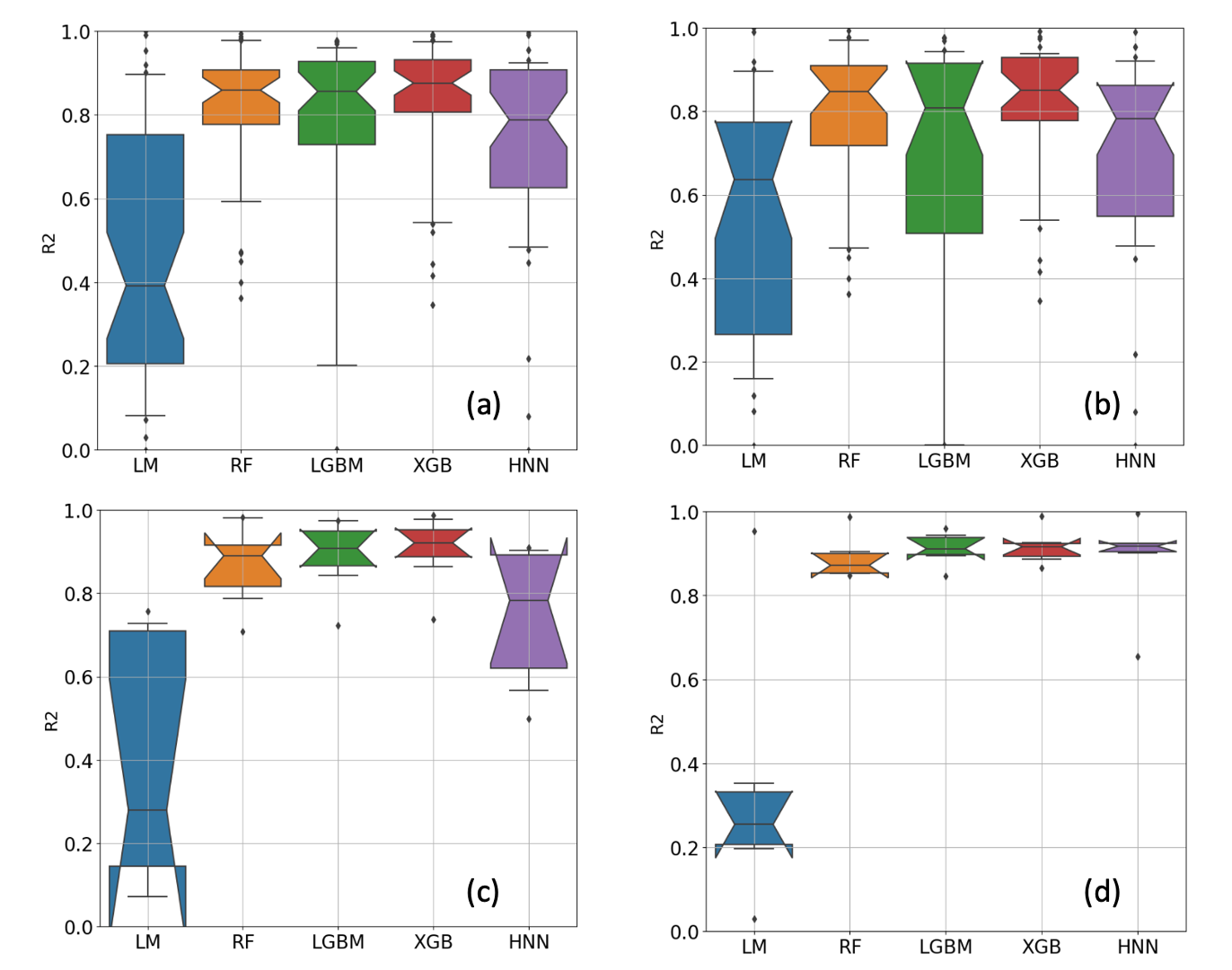}
\caption{
\label{pmlb_reg}
\footnotesize
The R2 score from different models on PMLB (Penn Machine Learning Benchmarks) regression datasets.
{\bf (a)} All datasets (46 datasets).
{\bf (b)} Datasets with number of variable $\in [0, 20)$ (32 datasets).
{\bf (c)} Datasets with number of variable  $\in [20, 40)$ (8 datasets).
{\bf (d)} Datasets with number of variable  $\in [40, \inf)$ (6 datasets).
}
\end{figure}

Table \ref{Tab:pmlb_reg_table2} and Figure \ref{pmlb_reg} report the comparison between HNN results and the machine learning methods on the PMLB regression datasets. We underline that HNN outperforms traditional machine learning methods and nearly matches the state-of-the-art. Furthermore, the relative performance of HNN improves with the number of variables, notably with HNN obtaining equivalent and marginally better performance even than XGB for the datasets with large number of features (see Figure \ref{pmlb_reg}(d)).

\subsection{Multivariate Time-series Data}\label{sec:expTS}

\begin{table*}[h!]
\center
\setlength\tabcolsep{6pt}
\begin{tabular}{c c|llll|llll}
\toprule
& & \multicolumn{4}{|c|}{solar-energy} & \multicolumn{4}{|c}{exchange-rates} \\
\midrule
& & \multicolumn{4}{|c}{Horizon (days)} & \multicolumn{4}{|c}{Horizon (days)} \\
Model & Metrics &         3 &    6 &   12 &   24 &    3 &    6 &   12 &   24          \\
\midrule
\midrule
LSTM-HNN & RSE &  $\textbf{0.190}^{*}$ & $\textbf{0.270}^{*}$ &  $\textbf{0.354}^{*}$ &  $\textbf{0.446}^{*}$ &  $\textbf{0.022}^{*}$ &  $\textbf{0.027}^{**}$ &  $\textbf{0.040}^{*}$ &  $\textbf{0.049}^{*}$ \\
       & CORR &  \textbf{0.981} &  $\textbf{0.964}^{*}$ &  $\textbf{0.942}^{*}$ &  $\textbf{0.902}^{**}$ &  $\textbf{0.976}^{***}$ &  $\textbf{0.968}^{**}$ &  $\textbf{0.956}^{*}$ &  $\textbf{0.938}^{*}$ \\
       \midrule
       \midrule
LSTM-MLP-HNN & RSE &  0.207 &  0.292 &  0.365 &  0.454 &  0.028 &  0.034 &  0.046 &  0.054 \\
       & CORR &  0.980 &  0.959 &  0.936 &  0.893 &  0.965 &  0.957 &  0.945 &  0.928 \\
       \midrule
LSTM-MLP-res & RSE &  0.245 &  0.340 &  0.409 &  0.501 &  0.031 &  0.035 &  0.052 &  0.059 \\
       & CORR &  0.972 &  0.944 &  0.905 &  0.898 &  0.850 &  0.829 &  0.835 &  0.828 \\
              \midrule
LSTM-MLP & RSE &  0.307 &  0.361 &  0.425 &  0.697 &  0.029 &  0.037 &  0.054 &  0.056 \\
       & CORR &  0.956 &  0.937 &  0.898 &  0.723 &  0.845 &  0.838 &  0.834 &  0.824 \\
\bottomrule
\end{tabular}
\vspace{0.5pt}
    \caption{Relative Standard Error (RSE) and CORR (correlation). The best-performing results in a given metric and horizon are highlighted in bold. In addition, a paired T-test has been performed, and the p-values for the LSTM-HNN against the second-best-performing model (i.e. the LSTM-MLP-res) in the given metrics and horizon are highlighted next to the best-performing results, where ${}^{*}$ denotes $1\%$ significance, ${}^{**}$ for $0.1\%$ and ${}^{***}$ for $0.001\%$ respectively.}
    \label{Tab:comparison table HNN}
\end{table*}
\begin{table*}[h]
\setlength\tabcolsep{6 pt}
\begin{center}
\begin{tabular}{c c|llll|llll}
\toprule
& & \multicolumn{4}{|c|}{solar-energy} & \multicolumn{4}{|c}{exchange-rates} \\
\midrule
& & \multicolumn{4}{|c}{Horizon (days)} & \multicolumn{4}{|c}{Horizon (days)} \\
Model & Metrics &         3 &    6 &   12 &   24 &    3 &    6 &   12 &   24          \\
\midrule\midrule
LSTM-HNN & RSE &  $0.190$ &  $0.270$ &  $0.354$ &  $0.446$ &  $0.022$ &  $0.027$ &  $0.040$ &  $0.049$ \\
       & CORR &  $0.981$ &  $0.964$ &  $0.942$ &  $0.902$ &  $0.976$ &  $0.968$ &  $0.956$ &  \textbf{0.938} \\
\midrule
\midrule
MTGNN & RSE &  $\textbf{0.177}^{*}$ &  $\textbf{0.234}^{**}$ &  $\textbf{0.310}^{*}$ &  $\textbf{0.427}^{*}$ &  0.019 &  0.025 &  0.034 &  0.045 \\
       & CORR &  \textbf{0.985} &  $\textbf{0.972}^{*}$ &  $\textbf{0.950}^{*}$ &  0.903 &  0.978 &  0.970 &  0.955 &  0.937 \\
       \midrule
TPA-LSTM & RSE &  0.180 &  0.234 &  0.323 &  0.438 &  $\textbf{0.017}^{*}$ &  \textbf{0.024} &  \textbf{0.034} & \textbf{ 0.044} \\
       & CORR &  0.985 &  0.974 &  0.948 & $ \textbf{0.908}^{*}$ &  $\textbf{0.979}^{*}$ &  0.970 &  \textbf{0.956} &  0.938 \\
       \midrule
LSTNet-skip 
& RSE &  0.184 &  0.255 &  0.325 &  0.464 &  0.022 &  0.028 &  0.035 &  0.044 \\
& CORR &  0.984 &  0.969 &  0.946 &  0.887 &  0.973 &  0.965 &  0.951 &  0.935 \\
      \midrule
RNN-GRU & RSE &  0.193 &  0.262 &  0.416 &  0.485 &  0.019 &  0.026 &  0.040 &  0.062 \\
       & CORR &  0.982 &  0.967 &  0.915 &  0.882 &  0.978 &  \textbf{0.971} &  0.953 &  0.922 \\
       \midrule
GP &  RSE &  0.225 &  0.328 &  0.520 &  0.797 &  0.023 &  0.027 &  0.039 &  0.058 \\
& CORR &  0.975 &  0.944 &  0.851 &  0.597 &  0.871 &  0.819 &  0.848 &  0.827 \\
       \midrule
VARMLP 
& RSE &  0.192 &  0.267 &  0.424 &  0.684 &  0.026 &  0.039 &  0.040 &  0.057 \\
& CORR &  0.982 &  0.965 &  0.905 &  0.714 &  0.860 &  0.872 &  0.828 &  0.767 \\
\midrule
AR & RSE &  0.243 &  0.379 &  0.591 &  0.869 &  0.022 &  0.027 &  0.035 &  0.044 \\
       & CORR &  0.971 &  0.926 &  0.810 &  0.531 &  0.973 &  0.965 &  0.952 &  0.935 \\   
\bottomrule
\end{tabular}
\end{center}
\vspace{0.5pt}
    \caption{Relative Standard Error and correlation. The best-performing results in a given metric and horizon are highlighted in bold. In addition, a paired T-test has been performed, and the p-values for the best-performing result against LSTM-HNN in the given metrics and horizon are highlighted next to the best-performing results, where ${}^{*}$ denotes $1\%$ significance, ${}^{**}$ for $0.1\%$ and ${}^{***}$ for $0.001\%$ respectively. 
    The absence of ${}^{*}$ indicates statistical equivalence between the best-performing and LSTM-HNN models.   
    When LSTM-HNN is the best-performing result, then the t-test is conversely performed against the second best-performing result. 
    }
    \label{Tab:comparison table}
\end{table*}
The HNN module can be used as a portable component along with different types of neural networks to manage various input data structures and downstream tasks. In this Section, we apply HNN to process dependency structures in time series modelling after temporal dependencies are handled through the LSTM architecture. We use two different datasets which have been extensively investigated in the multivariate time-series literature \cite{Wu2020ConnectingTD}: the solar-energy dataset from the National Renewable Energy Laboratory, which contains the solar-energy power output collected from 137 PV plants in Alabama State in 2007; and a financial dataset containing the daily exchange-rates rates of eight foreign countries including Australia, British, Canada, Switzerland, China, Japan, New Zealand, and Singapore in the period from 1990 to 2016 (see Table \ref{Tab:mtdata} in Appendix for further details).

Analogously with the tabular data, we first compare the outcomes of LSTM-HNN with those obtained with adapted MLP units. Specifically, LSTM units plus an MLP (LSTM-MLP); LSTM units plus an MLP with added residual connections to the final read-out layer (LSTM-MLP-res); and LSTM units plus a sparse MLP of the same layout as HNN without residual connections (LSTM-MLP-HNN). 
We then compare the LSTM-HNN results with traditional and state-of-the-art spatio-temporal models for multivariate time-series problems: auto-regressive model (AR) \cite{Zivot2003VectorAM}; a hybrid model that exploits both the power of MLP and auto-regressive modelling (VARMLP) \cite{Zhang2003TimeSF}; a Gaussian process (GP) \cite{Roberts2012GaussianPF}; a recurrent neural network with fully connected GRU hidden units (RNN-GRU) \cite{Wu2020ConnectingTD}; a LSTM recurrent neural network combined with a convolutional neural network (LSTNet) \cite{Lai2017ModelingLA}; a LSTM recurrent neural network with attention mechanism (TPA-LSTM) \cite{Shih2018TemporalPA}; and a graph neural network with temporal and graph convolution (MTGNN) \cite{Wu2020ConnectingTD}. 

We evaluate performances of the LSTM-HNN and compare them with the ones achieved by benchmark methodologies by forecasting the solar-energy power outputs and the exchange-rates values at different time horizons with performances measured in terms of relative standard error (RSE) and correlation (CORR) (see Table \ref{Tab:comparison table HNN}). We underline that LSTM-HNN significantly outperforms all MLP-based models. On solar-energy data, LSTM-HNN reduces RSE by 38\%, 25\%, 17\%, and 36\% from LSTM-MLP and 8\%, 7\%, 3\%, and 2\% from LSTM-MLP-res across four horizons. On exchange-rates data, LSTM-HNN reduces RSE by 23\%, 28\%, 26\%, and 13\% from LSTM-MLP and 19\%, 20\%, 14\%, and 10\% from LSTM-MLP-res across four horizons.

We also notice that the residual connections from each layer to the final read-out layer are effective both in the HNN architecture (i.e. LSTM-HNN outperforms LSTM-MLP-HNN) and within the MPL models (i.e. LSTM-MLP-res outperforms LSTM-MLP). 
In order to illustrate the significance of the gain, a paired t-test of LSTM-HNN against LSTM-MLP-res has been performed revealing that all differences are significant at 1\% or better with the only exception for the correlation at horizon 3 in the solar-energy output data. 

The comparison between the results for LSTM-HNN and the other benchmark models is reported in Table \ref{Tab:comparison table}.
Results reveal that LSTM-HNN consistently outperforms all three non-RNN-based methods (AR, VARMLP and GP) on both datasets. 
It also outperforms LSTNet-skip results. 
LSTM-HNN outperforms RNN-GRU for all datasets and horizons except for the correlation in the exchange rates at horizon 6 where it returns an equivalent result accordingly with the paired t-test that was conducted between LSTM-HNN and the best-performing model. 
LSTM-HNN is instead slightly outperformed by MTGNN in most results for solar-power and by TPA-LSTM in several results for exchange-rates. 
It must be however noticed that these are massive deep-learning models with a much larger number of parameters (respectively 1.5 and 2.5 times larger than LSTM-HNN for the solar-energy datasets and 10 and 26 times larger for the exchange-rates datasets, see Table \ref{Tab:param table HNN}).

\section{Conclusion}\label{sec:conclusion}
In this paper we introduce Homological Neural Networks (HNNs), a novel deep-learning architecture based on a higher-order network representation of multivariate data dependency structures. This architecture can be seen as a sparse MLP with extra residual connections and it can be applied in place of any fully-connected MLP unit in composite neural network models. We test the effectiveness of HNNs on tabular and time-series heterogeneous datasets. 
Results reveal that HNN, used either as a standalone model or as a modular unit within larger models, produces better results than MLP models with the same number of neurons and layers. We compare the performance of HNN with both fully-connected MLP, MLP sparsified with the HNN layered structure, and fully-connected MLP with additional residual connections and read-out unit. We design an experimental pipeline that verifies that the sparse higher-order homological layered structure on which HNN is built is the main element that eases the computational process. Indeed, we verify that the sparsified MLP with the HNN structure (MLP-HNN) over-performs all other MLP models. We also verify that the residual links between layers and the readout unit consistently improve HNN performances. Noticeably, although residual connections also improve fully-connected MLP performances, results are still inferior to the ones achieved by sparse MLP-HNN.  
We demonstrate that HNNs' performances are in line with state-of-the-art best-performing computational models, however, it must be considered that they have a much smaller number of parameters, and their processing architecture is easier to interpret.

In this paper, we build HNNs from TMFG networks computed on pure correlations. TMFG are very convenient chordal network representations that are computationally inexpensive and provide opportunities for dynamically self-adjusting neural network structures. Future research work on HNN will focus on developing an end-to-end dynamic model that addresses the temporal evolution of variable interdependencies. TMFG is only one instance of a large class of chordal higher-order information filtering networks \cite{Massara2019LearningCF} which can be used as priors to construct HNN units. The exploration of this larger class of possible representations is a natural expansion of the present HNN configuration and will be pursued in future studies.

\section*{Acknowledgements}

All the authors acknowledge the members of the University College London Financial Computing and Analytics Group for the fruitful discussions on foundational topics related to this work. All the authors acknowledge the ICML TAG-ML 2023 workshop organising committee and the reviewers for the useful comments that improved the quality of the paper. The author, T.A., acknowledges the financial support from ESRC (ES/K002309/1), EPSRC (EP/P031730/1) and EC (H2020-ICT-2018-2 825215).


\bibliography{icml2023/HNN_ref}
\bibliographystyle{icml2023}

\newpage
\appendix
\onecolumn
\section{Appendix}

\begin{table}[h!]
\center
\begin{tabular}{c|c|c|c}
\toprule
Dataset & {No. Features} & {No. Samples} & {Sample Rate} \\

\midrule
solar-energy & 137 & 52560 & 10 minutes \\
       \midrule
exchange-rates &  8 & 7588 & 1 day \\
\bottomrule
\end{tabular}

\vspace{0.5pt}
    \caption{Multivariate time-series dataset statistics, including the number of features, number of samples and sample rate in the solar-energy-energy and exchange-rates-rates datasets \cite{Wu2020ConnectingTD}. 
    }
    \label{Tab:mtdata}
\end{table}

\begin{table}[h!]
\center
\begin{tabular}{c|c|c}
\toprule
& {solar-energy} & {exchange-rates} \\

\midrule
LSTM-MLP & 452901 & 13011 \\
       \midrule
LSTM-MLP-HNN &  509208 & 13203 \\
       \midrule
LSTM-MLP-res &  509208 & 13203 \\
       \midrule
LSTM-HNN & 239061 & 12795\\
\bottomrule
\end{tabular}

\vspace{0.5pt}
    \caption{Number of parameters in each model in solar-energy and exchange-rates datasets, comparing the sparse LSTM-HNN with the fully connected models.}
    \label{Tab:param table HNN}
\end{table}

\begin{table}[h!]
\center
\begin{tabular}{c|c|c}
\toprule
& {solar-energy} & {exchange-rates} \\

\midrule
LSTM-skip & 337112 & 19478 \\
       \midrule
TPA-LSTM &  613987 & 132172 \\
       \midrule
MTGNN &  347665 & 337345 \\
       \midrule
LSTM-HNN & 239061 & 12795\\
\bottomrule
\end{tabular}

\vspace{0.5pt}
    \caption{Number of parameters in each model in solar-energy and exchange-rates datasets, comparing the LSTM-HNN with respect to  state-of-art models.}
    \label{Tab: param table}
\end{table}

\end{document}